\documentclass{article} % For LaTeX2e
\usepackage{iclr2023_conference,times}

\usepackage{amsmath,amsfonts,bm}

\def\eqref#1{equation~\ref{#1}}
\def\1{\bm{1}}

\DeclareMathAlphabet{\mathsfit}{\encodingdefault}{\sfdefault}{m}{sl}
\SetMathAlphabet{\mathsfit}{bold}{\encodingdefault}{\sfdefault}{bx}{n}

\usepackage{comment}
\usepackage{hyperref}
\usepackage{url}
\usepackage{subcaption}
\usepackage{graphicx}
\usepackage{bbm}
\usepackage{wrapfig}
\usepackage{listings}
\usepackage[font=small,labelfont=bf]{caption}
\title{Bird distribution modelling using remote sensing and citizen science data}

\author{Mélisande Teng \\%,\thanks{ Use footnote for providing further information
Mila, Université de Montréal\\
\texttt{tengmeli@mila.quebec} \\
\And
Amna Elmustafa \\
Mila, Stanford University \\
\texttt{amna97@stanford.edu} \\
\And
Benjamin Akera\\
Mila, McGill University \\
\texttt{akeraben@mila.quebec} \\
\And
Hugo Larochelle\\
Mila, Google Research, Brain Team\\
\texttt{hugolarochelle@google.com} \\
\And
David Rolnick \\
Mila, McGill University\\
\texttt{drolnick@cs.mcgill.ca}
}

 \iclrfinalcopy % Uncomment for camera-ready version, but NOT for submission.
\begin{document}

\maketitle

\begin{abstract}
Climate change is a major driver of biodiversity loss, changing the geographic range and abundance of many species. However, there remain significant knowledge gaps about the distribution of species, due principally to the amount of effort and expertise required for traditional field monitoring. We propose an approach leveraging computer vision to improve species distribution modelling, combining the wide availability of remote sensing data with sparse on-ground citizen science data.% platforms.
We introduce a novel task and dataset for mapping US bird species to their habitats by predicting species encounter rates from satellite images, along with baseline models which demonstrate the power of our approach. Our methods open up possibilities for scalably modelling ecosystems properties worldwide.

\end{abstract}

\section{Introduction}\label{sec:intro}
Climate change presents significant threats to global biodiversity, both directly and by compounding the effects of other anthropogenic stressors, such as habitat loss, pollution, and introduced species. Biodiversity loss in turn also impacts ecosystem services necessary to ensure food, water, and human health and well-being \citep{marselle2021pathways}.

 It is crucial to understand the changing distributions of species globally to inform policy decisions, for example in shaping land use and land conservation choices. However, traditional methods for species distribution models (SDMs) generally focus either on narrow sets of species or narrow geographical areas. This is in part due to the methods' computational cost, the insufficient ability of models to account for complex relationships between different variables, and the type of data used, whose availability can be limited \citep{phillips2005brief, newbold2010applications}.

Machine learning algorithms for remote sensing have increasingly seen wide applicability across sustainability-related domains  (see e.g.~\cite{rolf2021mosaiks}) and have been suggested as a promising tool for SDMs \citep{beery2021sdm_review}. Moreover, the surge of data collection on citizen science platforms along with improved data quality validation processes in recent years offers tremendous opportunity for scientific research, as species observation records from these sources can cover a larger temporal and geographic extent at a finer resolution and at lower cost than traditional sampling methods. Indeed, citizen science data and remotely sensed ecosystem attributes such as vegetation indices have been shown to improve the performance of models, especially for less widespread species \citep{arenas2022effects}. %However, in these models the input data is often only available at a coarse resolution (typically 1 km), which is insufficient to understand fine-grained habitats.
Recently, the GeoLifeCLEF challenge \citep{lorieul2022overview} was introduced with the goal of directly predicting plant and animal abundance from aerial images at 1$m$ resolution, using deep computer vision. However, the GeoLifeCLEF benchmark has proven extremely challenging, potentially since only one species is associated with each location.

We propose to use remote sensing to infer the joint distribution of many species for a given location, using publicly available citizen science observation records as ground truth. Our approach leverages the fact that a species' presence or absence at a location depends on the ecosystem present there, and therefore the abundances of different species are highly correlated. Specifically, we predict the encounter rates of 684 bird species at sites across the continental USA. We focus on birds due to the availability of high-quality bird observations and the ecological importance of birds, which play a vital role in almost every terrestrial ecosystem and are threatened significantly by climate change \citep{rodenhouse2008potential, stephens2016consistent}. Our contributions include: (i) framing the task of predicting bird species encounter rates at a specific location using remote sensing data, (ii) building a dataset for this task, obtained from publicly available bird observation and satellite data sources, and (iii) showing the efficacy of baseline deep computer vision methods for this task.

\paragraph{Problem definition.}
We consider presence-absence bird sighting records from the citizen science database eBird \citep{eBird:HCLN}. Each location (termed a \emph{hotspot}) in the eBird database is associated to a number of \textit{complete checklists}  containing all species a birdwatcher was able to observe at a specific date and time.
If $h$ is a hotspot, and $s_1,\ldots,s_n$ the species of interest, then our goal is to build a machine learning model that takes as input a satellite image of $h$ (and optionally other data) and predicts the vector $\mathbf{y^h} = (y_{s_1}^h, ..., y_{s_n}^h)$, where $y_s^h$ is the number of complete checklists reporting species $s$ at $h$ divided by the total number of complete checklists at $h$.
%\begin{equation}
%    y_s^h = \dfrac{\text{number of complete checklists reporting species $s$ at $h$}}{\text{total number of complete checklists at $h$}}
%\label{eq:ebird_target}
%\end{equation} 
This ratio $y_s^h$ can be understood as an \emph{encounter rate}, the probability for a visitor to observe a species if they visit the hotspot. We aim at jointly predicting this quantity for all bird species in the region. Thus, our task can be considered as a supervised multi-output regression problem. We consider encounter rates as our target variable because they are a widely used measure in species distribution modeling, and are indeed used extensively on the eBird platform (in the form of ``hotspot bar charts'') in order to summarize the species in a given location for birdwatchers and ornithologists. 

\section{Dataset}

 We introduce a dataset for the task defined in \ref{sec:intro} with the continental USA as our region of interest. Bird distributions vary seasonally; we consider specifically the month of June, representing the breeding period for most birds in the region, and making it possible to neglect migratory effects. Our dataset will be released publicly upon acceptance. 

\textbf{Species observation data.} We extracted all eBird complete checklists recorded in June from 2000 to 2021 in the continental USA. We filtered out hotspots with fewer than 5 complete checklists recorded in June to ensure more meaningful encounter rates, resulting in 8439 hotspots. We included all regularly occurring species in the region, as denoted by ABA Codes 1 and 2 \citep{american2022aba}, omitting species found only in Hawaii and Alaska, as well as one Code-2 seabird species with rare oceanic observations, leaving a total of 684 species. 
We then computed the encounter rates for each hotspot and corrected for vagrants (species seen in hotspots outside of their geographical range) by using range maps from eBird to set the target encounter rates to zero in these hotspots. We aggregate the species observations in June over twenty years, only considering seasonal change in distributions. We leave annual temporal change for future iterations of this work.  %In the first iteration of our models, we do not consider temporal change over We are interested in the seasonality of species distribution rather than annual change, so for the first iteration of our models we aggregate the species observations over 20 years. In future work 

\textbf{Remote Sensing Imagery.} For each hotspot, we extracted RGB and NIR bands (reflectance measurements) from Sentinel-2 satellite tiles covering the entirety of a square of about 5 km$^2$ centered around the hotspots. The images have a resolution of 10 $m$ for the considered bands. We extracted images in June with cloud coverage of at most 10 \%, keeping the least cloudy and most recent image of those in the period 2016-2020 if it covered the entire 5 km$^2$ region or composing a mosaic with the extracted images otherwise, in order to minimize seams in the images of our dataset. We associate one image per hotspot for years from 2016-2020,, considering that a more recent satellite image would be more representative of our species data, since there are more checklists in recent years compared to earlier years.

\textbf{Environmental data.}  We extracted 19 bioclimatic and 8 pedologic variables as rasters from WorldClim 1.4 and SoilGrids in the same fashion as the GeoLifeCLEF 2020 dataset \citep{cole2020geolifedata} for each hotspot. Additionally, we extracted land cover data patches from the ESRI Land Use Land Cover 10m-resolution maps derived from Sentinel-2 \citep{esri2020landcover}.

\textbf{Train-test split} In order to account for spatial auto-correlation and overfitting that can arise from random splits of geospatial data \citep{roberts2017cross}, we first clustered hotspots such that the minimum distance between clusters is 5 $km$ and randomly assigned them to train, validation and test splits in a 70\%, 20\%, 10\% proportion.

\section{Methodology}\label{ref:method}
In this section, we detail baselines for this task as well as directions for further improvement.

\subsection{Environmental baseline} A first naive baseline is the mean encounter rates over the training set for each species. Following the GeoLifeCLEF challenge, we  also propose an environmental baseline, using Gradient Boosted Regression Trees on the bioclimatic and pedological variables extracted at each of the hotspots.

\subsection{CNN baseline}
We propose a first CNN model with ResNet-18 \citep{he2016deep} architecture. We initialize the network with ImageNet pretrained weights. Since ImageNet has only RGB bands, the initialization of the first layer for the NIR band is performed by sampling from a normal distribution parameterized by the mean and standard deviation of the layer's weights for the other bands.  

 We consider a region of interest of 640 m$^2$, center-cropping the satellite patches to size $64\times64$ around the hotspot and normalizing the bands with our training set statistics. We use random vertical and horizontal flipping for data augmentation and train the model with cross entropy $\mathcal{L}_{CE}= \frac{1}{N_h} \sum_{h} \mathcal{L}_h = \frac{1}{N_h} \sum_{h} \sum_{s \text{(species)}} -y^s_h \log(\hat{y}^s_h) - (1-y^s_h)\log(1-\hat{y}^s_h) $ where $N_h$ is the number of hotspots $h$, $y$ the predictions of the model and $\hat{y}$ the ground truth encounter rates.
 
We add the bioclimatic and pedological data by normalizing it variable-wise with training set statistics, aligning it to the satellite images' resolution and stacking the corresponding patches to the images. We add the landcover data following the same procedure. 

\subsection{Including geographical information}
We explore different methods for making the model spatially explicit to account for the geographical ranges of some species. For example, Eastern and Western wood pewees have the same habitat but are not found in the same geographical regions, and thus do not co-occur. A model blind to geographical information might predict both species in the same place, which is undesirable when producing maps for conservation planning, and could also impede training.

\textbf{Location encoder (LE).} We train a location encoder taking as input latitude and longitude information jointly with the image encoder. We use the same architecture as the location encoder of \cite{mac2019presence} (cf.~\ref{sup:location encoder}), removing the final classification layer, and instead concatenating the obtained features to those obtained from the ResNet image encoder, before passing them to a fully-connected layer to obtain the predicted encounter rates.

\textbf{Hard-masking with range maps (RM).}
Some species have known geographical range. In this approach, we use range maps available via eBird which are updated regularly as binary masks on the predictions to zero out the encounter rates of species in regions where they cannot be found. If the range map is not available for a species, the predictions are left untouched. 

\textbf{Soft-masking (SM).} Here, we compute proxies for range maps from the training set data, since range maps (a) are available only for certain species and (b) do not capture finer gradations of abundance across regions.
We define a regional ``correction'' factor $c_R$, which reflects the relative prevalence of species in region $R$ compared to the prevalence in the whole geographical area $W$. In our case, $R$ varies across US states, while $W$ is the continental US. For each species $s$, we set:
\begin{equation}
c_s^R = \frac{\sum_{l \in C_R} \mathbbm{1}_{s\in l}}{|C_R|} / \frac{\sum_{l \in C_W} \mathbbm{1}_{s\in l}}{|C_W|}
\end{equation}
where $C_R$ and $C_W$ are the set of observations in $R$ and $W$ resp., and $s \in l$ denotes that $s$ has been reported in $l$. For a hotspot $h$ in $R$, the final prediction is 
\lstinline{clip}$_{[0,1]} (c_R . y_h)$. See \ref{sub:softmask} for derivation.

\subsection{Including hotspot metadata} 
Additionally, we consider a weighted loss (WL) with weights as a function of the number of checklists $n_h$ at each hotspot, and give more importance to more visited hotspots since the data at such hotspots is likely more reliable by virtue of more checklists. 
The contribution of each hotspot to the loss is 
 $f(n_h)\mathcal{L}_h $. For $f$, we consider the identity, $\log$ and square root functions.

\subsection{Evaluation} 

Beyond optimization for common regression metrics (MSE, MAE), it is desirable for the predicted most likely species in a given hotspot to coincide with those that have most frequently been observed. We therefore report top-10 and top-30 accuracies, representing for $k=10$ and $k=30$ the number of species present in both the top $k$ predicted species and the top $k$ observed species, divided by $k$. However, the number of species observed varies considerably across  hotspots (cf. \ref{sup:training-distrib}), unlike in e.g.~the GeoLifeCLEF challenge for which there is only one ground truth species for each hotspot. We therefore also define an \textit{adaptive top-$k$ accuracy} as the top-$k$ value where $k$ is the total number of species observed at that hotspot. We also analyse the predictions of the model by looking at precision and recall per species across hotspots and provide further details in \ref{sup:precrec}.

\section{Experiments}
\begin{wrapfigure}[15]{h}{0.45\textwidth}
\includegraphics[width=6.7cm]{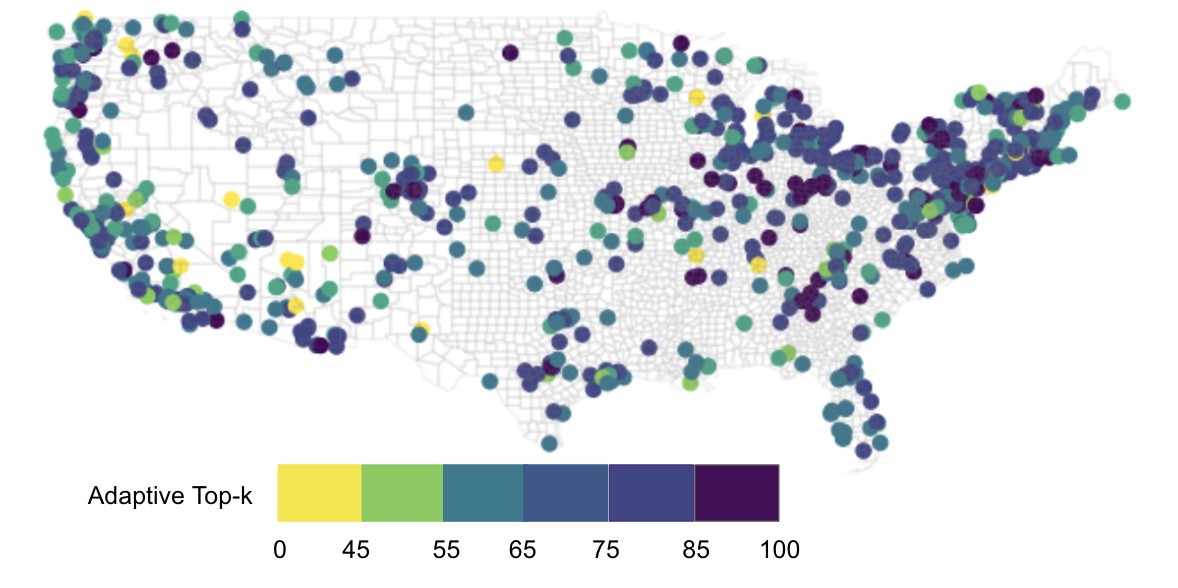}\label{fig:plot}
\caption{Test set hotspots colored by adaptive top-k performance for the RGBNIR + Env + range maps masking model.}
\centering
\end{wrapfigure}
Numerical results on the test set for the models presented in Section \ref{ref:method} are provided in Table \ref{tab:results} and validation performance is reported in \ref{sup:val_perf}. All models were trained with batch size 64, Adam optimizer, initial learning rate $0.0003$, with a scheduler that decreases the rate based on validation loss.

While the CNN baseline with RGB and NIR bands does not outperform the environmental baseline on all metrics, combining satellite and environmental data outperforms the latter, particularly on the top-k metric. This highlights the importance of remote sensing data for our task. Adding landcover data did not improve the model, likely due to the coarseness of the 10 classes, so it was not used for further experiments.
While adding geographical information through RM masking improves the performance of the model, the LE and the SM methods did not, perhaps because of the simplicity of the architecture of the former and the coarse region unit (US states) considered for the latter. We plan to explore other location encoding methods, such as space2vec \citep{Mai2020Multi-Scale} and consider smaller units for the soft-masking. Fig.~1 shows the geographical distribution of performance on the test set of the hard-masking with RM model and we provide further examples of predictions in \ref{sup:predictions}. Species-wise precision and recall are reported in \ref{sup:precrec}. We find that range-restricted species, along with very common species, are the best performing species according to recall.
Interestingly, the weighted loss yields performance similar to that of regular cross-entropy loss. The best performance was achieved with the log function. This suggests that checklists from less frequented hotspots (e.g.~with fewer than 5 complete checklists) could be included in future iterations without compromising performance.
\begin{table}[ht]
\centering
\scriptsize
\caption[1]{Results for the proposed models on the test set. %\protect\footnotemark.
``Env'' indicates the use of environmental variables. All MSE and MAE scores are reported $\times 10^3$ and $\times 10^2$ respectively. ``Top-$k$'' denotes the adaptive top-$k$ accuracy.}
\begin{tabular}{|l || c |c  | c |c |c|} 

 \hline
 \textbf{Method} &\textbf{MSE} [$10^{-3}$]&\textbf{MAE} [$10^{-2}$]&\textbf{Top-$k$} &\textbf{Top-30}&\textbf{Top-10}\\
  % & Test &Val & Test\\ [0.5ex] 
\hline
 Mean encounter rate  & $7.18$ &  $2.91$ & 51.46 &43.91 &26.45\\ 
 
 Env baseline & \strut \hbox{4.83$\pm 0.00$} & $2.05\pm0.00$  &$68.86\pm0.01 $  &$62.3\pm0.03 $  &$43.11\pm0.06 $\\
 \hline
 RGBNIR & \strut \hbox{5.39$\pm 0.00$} & $1.98\pm0.00$ & 67.0 $\pm0.02 $  & 60.1  $\pm0.17 $& 41.83  $\pm0.30 $ \\
 
 RGBNIR + Env & 4.57$\pm 0.02$ & $1.78 \pm 0.01$  & 72.91 $\pm 0.19$ & 67.0  $\pm0.00 $& 48.0  $\pm0.00 $\\
%RGBNIR + Env + Landuse & 4.567$\times10^{-3}\pm 1.8\times10^{-6}$ & $1.78 \times10^{-2}\pm8.56\times10^{-5}$  & 72.91 $\pm 1.95\times10^{-3}$\\

 \hline
   RGBNIR + Env + LE &
   $4.49\pm 0.05$ &1.73$\pm$ 0.01 &  $72.01 \pm 0.08$  &65.00  $\pm0.00 $& 44.50  $\pm0.70 $ \\
  \textbf{RGBNIR + Env + RM }& \textbf{4.45$\pm$ 0.05} &1.75$\pm$ 0.01 &  $\mathbf{73.38 \pm 0.08}$&\textbf{67.68  $\pm$0.11 } &\textbf{48.15}  $\pm$\textbf{0.20 } \\
   RGBNIR + Env + SM &\strut \hbox{13.3$\pm 0.39$}&	\strut \hbox{2.74
  $\pm0.08$}&	\strut \hbox{65.0 $\pm 1.39$}& 54.7  $\pm 2.3 $& 31.7  $\pm1.04 $\\
  % RGBNIR + Env + LE +RM& $4.48\times10^{-3}\pm 5.32\times10^{-5}$ &$1.71\times10^{-2}\pm 1.73\times10^{-4}$ &  $72.45 \pm 2.53\times10^{-2}$  \\
  \hline
  RGBNIR + Env + RM + WL & 4.76 $\pm$0.00 & \textbf{1.60 $\pm$ 0.00} & 72.45$ \pm 0.02$ & $65.90 \pm 0.02$ & $46.30 \pm 0.006 $\\
  % [1ex] 
 \hline
\end{tabular}

\label{tab:results}
\end{table}
\vspace{-5pt}
\section{Conclusion}
We introduce a new dataset built from publicly available data to leverage the combination of satellite imagery and presence-absence citizen science data for joint species distribution modelling. We are currently extending our approach and dataset to other taxonomic groups and regions. In East Africa, for example, sparse data poses additional challenges but also underscores the importance of scalable remote sensing approaches to supplement scarce on-the-ground data. We hope to integrate our algorithm into eBird's existing tool that lists the ``likely species'' in a given area. This tool currently relies on past checklists and is therefore only available in well-monitored locations. By approximating species distributions at poorly monitored locations using remote sensing to evaluate changing land use patterns and habitat, we anticipate significantly expanding the ability of ornithologists to rapidly estimate biodiversity across the world.
%add future work in there

\bibliography{iclr2023_conference}

\begin{thebibliography}{17}
\providecommand{\natexlab}[1]{#1}
\providecommand{\url}[1]{\texttt{#1}}
\expandafter\ifx\csname urlstyle\endcsname\relax
  \providecommand{\doi}[1]{doi: #1}\else
  \providecommand{\doi}{doi: \begingroup \urlstyle{rm}\Url}\fi

\bibitem[ABA(2022)]{american2022aba}
ABA.
\newblock \emph{{American Birding Association} checklist: Birds of the
  continental {United States and Canada}}.
\newblock 2022.

\bibitem[Arenas-Castro et~al.(2022)Arenas-Castro, Regos, Martins, Honrado, and
  Alonso]{arenas2022effects}
Salvador Arenas-Castro, Adri{\'a}n Regos, Ivone Martins, Jo{\~a}o Honrado, and
  Joaquim Alonso.
\newblock Effects of input data sources on species distribution model
  predictions across species with different distributional ranges.
\newblock \emph{Journal of Biogeography}, 49\penalty0 (7):\penalty0 1299--1312,
  2022.

\bibitem[Beery et~al.(2021)Beery, Cole, Parker, Perona, and
  Winner]{beery2021sdm_review}
Sara Beery, Elijah Cole, Joseph Parker, Pietro Perona, and Kevin Winner.
\newblock Species distribution modeling for machine learning practitioners: A
  review.
\newblock In \emph{ACM SIGCAS Conference on Computing and Sustainable
  Societies}, pp.\  329--348, 2021.

\bibitem[Cole et~al.(2020)Cole, Deneu, Lorieul, Servajean, Botella, Morris,
  Jojic, Bonnet, and Joly]{cole2020geolifedata}
Elijah Cole, Benjamin Deneu, Titouan Lorieul, Maximilien Servajean, Christophe
  Botella, Dan Morris, Nebojsa Jojic, Pierre Bonnet, and Alexis Joly.
\newblock The {GeoLifeCLEF} 2020 dataset.
\newblock \emph{Preprint arXiv:2004.04192}, 2020.

\bibitem[Esri(2020)]{esri2020landcover}
Esri.
\newblock Esri land cover 2020.
\newblock \url{https://livingatlas.arcgis.com/landcover/}, 2020.
\newblock Accessed: 2022-12-06.

\bibitem[He et~al.(2016)He, Zhang, Ren, and Sun]{he2016deep}
Kaiming He, Xiangyu Zhang, Shaoqing Ren, and Jian Sun.
\newblock Deep residual learning for image recognition.
\newblock In \emph{Proceedings of the IEEE conference on computer vision and
  pattern recognition}, pp.\  770--778, 2016.

\bibitem[Kelling et~al.(2013)Kelling, Gerbracht, Fink, Lagoze, Wong, Yu,
  Damoulas, and Gomes]{eBird:HCLN}
Steve Kelling, Jeff Gerbracht, Daniel Fink, Carl Lagoze, Weng-Keen Wong, Jun
  Yu, Theo Damoulas, and Carla Gomes.
\newblock {eBird}: A human/computer learning network for biodiversity
  conservation and research.
\newblock \emph{AI Magazine}, 34, 03 2013.

\bibitem[Lorieul et~al.(2022)Lorieul, Cole, Deneu, Servajean, Bonnet, and
  Joly]{lorieul2022overview}
Titouan Lorieul, Elijah Cole, Benjamin Deneu, Maximilien Servajean, Pierre
  Bonnet, and Alexis Joly.
\newblock Overview of {GeoLifeCLEF} 2022: Predicting species presence from
  multi-modal remote sensing, bioclimatic and pedologic data.
\newblock In \emph{CLEF 2022-Conference and Labs of the Evaluation Forum},
  volume 3180, pp.\  1940--1956, 2022.

\bibitem[Mac~Aodha et~al.(2019)Mac~Aodha, Cole, and Perona]{mac2019presence}
Oisin Mac~Aodha, Elijah Cole, and Pietro Perona.
\newblock Presence-only geographical priors for fine-grained image
  classification.
\newblock In \emph{Proceedings of the IEEE/CVF International Conference on
  Computer Vision}, pp.\  9596--9606, 2019.

\bibitem[Mai et~al.(2020)Mai, Janowicz, Yan, Zhu, Cai, and
  Lao]{Mai2020Multi-Scale}
Gengchen Mai, Krzysztof Janowicz, Bo~Yan, Rui Zhu, Ling Cai, and Ni~Lao.
\newblock Multi-scale representation learning for spatial feature distributions
  using grid cells.
\newblock In \emph{International Conference on Learning Representations}, 2020.

\bibitem[Marselle et~al.(2021)Marselle, Hartig, Cox, De~Bell, Knapp, Lindley,
  Triguero-Mas, B{\"o}hning-Gaese, Braubach, Cook,
  et~al.]{marselle2021pathways}
Melissa~R Marselle, Terry Hartig, Daniel~TC Cox, Si{\^a}n De~Bell, Sonja Knapp,
  Sarah Lindley, Margarita Triguero-Mas, Katrin B{\"o}hning-Gaese, Matthias
  Braubach, Penny~A Cook, et~al.
\newblock Pathways linking biodiversity to human health: A conceptual
  framework.
\newblock \emph{Environment International}, 150:\penalty0 106420, 2021.

\bibitem[Newbold(2010)]{newbold2010applications}
Tim Newbold.
\newblock Applications and limitations of museum data for conservation and
  ecology, with particular attention to species distribution models.
\newblock \emph{Progress in physical geography}, 34\penalty0 (1):\penalty0
  3--22, 2010.

\bibitem[Phillips et~al.(2005)]{phillips2005brief}
Steven~J Phillips et~al.
\newblock A brief tutorial on {M}axent.
\newblock \emph{AT\&T Research}, 190\penalty0 (4):\penalty0 231--259, 2005.

\bibitem[Roberts et~al.(2017)Roberts, Bahn, Ciuti, Boyce, Elith,
  Guillera-Arroita, Hauenstein, Lahoz-Monfort, Schr{\"o}der, Thuiller,
  et~al.]{roberts2017cross}
David~R Roberts, Volker Bahn, Simone Ciuti, Mark~S Boyce, Jane Elith, Gurutzeta
  Guillera-Arroita, Severin Hauenstein, Jos{\'e}~J Lahoz-Monfort, Boris
  Schr{\"o}der, Wilfried Thuiller, et~al.
\newblock Cross-validation strategies for data with temporal, spatial,
  hierarchical, or phylogenetic structure.
\newblock \emph{Ecography}, 40\penalty0 (8):\penalty0 913--929, 2017.

\bibitem[Rodenhouse et~al.(2008)Rodenhouse, Matthews, McFarland, Lambert,
  Iverson, Prasad, Sillett, and Holmes]{rodenhouse2008potential}
Nicholas~L Rodenhouse, SN~Matthews, KP~McFarland, JD~Lambert, LR~Iverson,
  A~Prasad, T~Scott Sillett, and Richard~T Holmes.
\newblock Potential effects of climate change on birds of the {N}ortheast.
\newblock \emph{Mitigation and adaptation strategies for global change},
  13:\penalty0 517--540, 2008.

\bibitem[Rolf et~al.(2021)Rolf, Proctor, Carleton, Bolliger, Shankar, Ishihara,
  Recht, and Hsiang]{rolf2021mosaiks}
Esther Rolf, Jonathan Proctor, Tamma Carleton, Ian Bolliger, Vaishaal Shankar,
  Miyabi Ishihara, Benjamin Recht, and Solomon Hsiang.
\newblock A generalizable and accessible approach to machine learning with
  global satellite imagery.
\newblock \emph{Nature communications}, 12\penalty0 (1):\penalty0 1--11, 2021.

\bibitem[Stephens et~al.(2016)Stephens, Mason, Green, Gregory, Sauer, Alison,
  Aunins, Brotons, Butchart, Campedelli, et~al.]{stephens2016consistent}
Philip~A Stephens, Lucy~R Mason, Rhys~E Green, Richard~D Gregory, John~R Sauer,
  Jamie Alison, Ainars Aunins, Llu{\'\i}s Brotons, Stuart~HM Butchart, Tommaso
  Campedelli, et~al.
\newblock Consistent response of bird populations to climate change on two
  continents.
\newblock \emph{Science}, 352\penalty0 (6281):\penalty0 84--87, 2016.

\end{thebibliography}
\bibliographystyle{iclr2023_conference}

\appendix

\section{Appendix}
\subsection{Training set distribution}\label{sup:training-distrib}
\begin{figure}[h]
\centering
\includegraphics[width=0.65\textwidth]{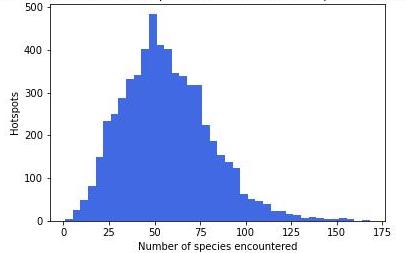}
\caption{Distribution of the number of species encountered in training set hotspots.}
\label{fig:dist_hist}
\end{figure}

A characteristic of the presence-absence data considered for this task is the zero-inflated nature of the targets. Indeed, while 684 species are considered, hotspots in our training set have 57 species with non-zero encounter rate on average. Fig. \ref{fig:dist_hist} shows the distribution of number of species encountered.

\subsection{Location encoder}\label{sup:location encoder}
\begin{figure}[h]
\centering
\includegraphics[width=0.5\textwidth]{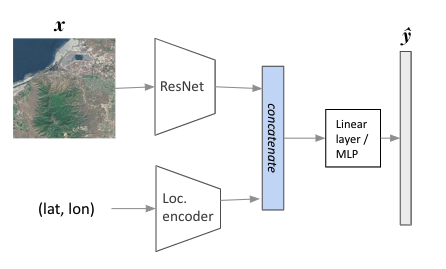}
\caption{Schematic view of the model including location information through a separate encoder.}
\label{fig:loc_encoder}
\end{figure}

We include latitude-longitude information through a separate encoder as shown in  Fig. \ref{fig:loc_encoder}.
We follow the implementation for \cite{mac2019presence} for the location encoder.
We first map each spatial coordinate $x^l$ (latitude and longitude) to $[\cos(\pi x^l), \sin(\pi x^l)]$. We end up with a input vector of size 4,  which is passed through an initial fully connected layer, followed by 4 residual blocks, each consisting of two fully connected layers with
a dropout layer in between.

\subsection{Derivation of the soft masking factor}\label{sub:softmask}
In this section we explain further our choice of the soft masking factor which draws inspiration from  the work of \cite{mac2019presence}. The underlying assumption is that while satellite images are informative about habitat, location also encodes geography, and both are important for designing a spatially-explicit model in our setting. To encode location here, we take $y$ as the target, $x$ as the input image and $l$ as the location in a region $R$. Predictions conditioned on both $x$ and $l$ can be expressed using Bayes rule:
\begin{equation}
p\left( y| x,l\right) =\dfrac{p\left( y| x\right) \ast p\left( x\right) \ast p\left( l| y,x\right) }{p\left( x,l\right) }
\end{equation}
Under the assumption that $l$ and $x$ are independent and also independent w.r.t $y$, we can approximate the formula as follows, where  $p\left( y| x\right)$ is the output of the CNN and $c^R$ is the correction factor. In fact, we account for geography at the level of region R, meaning that predictions in all location within R will be corrected with the same correction factor. 
\begin{equation}
 p\left( y| x,l\right) =\dfrac{p\left( y| x\right) \ast p\left( l| y\right) }{p\left( l\right) }
\end{equation}

\begin{equation}
c^R = \dfrac{p\left( l| y\right) }{p\left( l\right) }
\end{equation}
which is interpreted as the proportion of checklists in $R$ reporting y over the proportion of checklists anywhere  reporting $y$. This can be expressed in terms of counts for each species $s$ as :
\begin{equation}
c_s^R = \frac{\sum_{l \in C_R} \mathbbm{1}_{s\in l}}{|C_R|} / \frac{\sum_{l \in C_W} \mathbbm{1}_{s\in l}}{|C_W|}
\end{equation}
Note that although this factor is easy to calculate from the training data, and allow scalability  to sparse data regimes, this factor can results in predictions values greater than $1$, we propose to handle this by clipping the values to $1$. However, we would like to think in the future about better options. Moreover, the assumption of the independence of $l$ and $x$  is very strong, because satellite images are inherently tied to the location they correspond to. We suspect this may be a reason for the  degradation in performance in table \ref{tab:results} when using this factor. We aim to explore more about cases when this assumption doesn't hold. 
\subsection{Validation results}\label{sup:val_perf}
We provide results on the validation set of the proposed baseline models in Table \ref{tab:val_results}. 
\begin{table}[ht]
\scriptsize
\caption[1]{Results for the proposed models on the validation set.
The ranking of models is the same as on the test set. MSE and MAE are reported $\times 10^3$ and $\times 10^2$ respectively. }
\begin{tabular}{|l || c |c  | c |c|c|} 

 \hline
 \textbf{ Method} &\textbf{MSE} [$10^{-3}$]&\textbf{MAE}[$10^{-2}$]&\textbf{Top-$k$} &\textbf{Top-30}&\textbf{Top-10}\\
  % & Test &Val & Test\\ [0.5ex] 
\hline
 Mean ER  & $7.393$ &  $2.944$ & 50.75& $43.91$&$26.45$\\ 
 
 Env baseline & \strut \hbox{$4.93\pm0.0 $} & $2.08\pm0.0$  &$68.60 \pm 0.00$   &$61.64 \pm 0.01$&$42.41 \pm 0.01$ \\
 \hline
  RGBNIR & \strut \hbox{5.5$\pm 0.05$} & $1.99\pm0.00$ & 66.9 $\pm0.30 $&59.7  $\pm0.30 $&41.83  $\pm0.20 $ \\
 
 RGBNIR + Env & \strut \hbox{4.56$\pm0.02$}&	\strut \hbox{1.7$\pm 0.0$}&	\strut \hbox{73.0$\pm 0.00$} & 66.0 $\pm0.00 $&48.0  $\pm0.00 $\\

  RGBNIR + Env + LE &  \strut \hbox{4.9$\pm 0.04$}&	\strut \hbox{1.7$\pm 0.01$}&	\strut \hbox{72.5$\pm 0.00$}& 64.5  $\pm0.70 $&44.0  $\pm0.00 $\\
 
  RGBNIR + Env + RM &\strut \hbox{4.6$\pm 0.02$}&	\strut \hbox{1.8
  $\pm0.00$}&	\strut \hbox{73.3$\pm 0.00$}& 66.6  $\pm0.50 $&48.2  $\pm0.30 $\\

  RGBNIR + Env + SM &\strut \hbox{13.6$\pm 0.45$}&	\strut \hbox{2.78
  $\pm0.11$}&	\strut \hbox{64.5$\pm 1.56$}& 54.2  $\pm 1.75 $& 31.0  $\pm1.41 $\\

  \hline
  RGBNIR + Env + RM + WL & 4.76 $\pm 0.00$ & 1.77 $\pm 0.00$  & $72.4 \pm 0.01$ & 65.4 $\pm 0.02$ & 46.01 $\pm$ 0.02  \\
  % [1ex] 
 \hline
\end{tabular}

\label{tab:val_results}

\end{table}

\subsection{Top 10 Species predictions at different hotspots}\label{sup:predictions}
All predictions reported at those obtained with the RGBNIR + Env + RM + WL model.
We sampled hotspots from our test set and evaluated the model's performance in predicting the top 10 bird species at each location. We selected hotspots from geographically distinct regions, featuring different land cover types, including the Pacific Coast (California), Northeast (Maine) and Central US (Ohio). The following figures illustrate the results from these regions.

%####california

\begin{figure}[ht]
    \centering
    \includegraphics[width=\textwidth,height=0.9\textheight,keepaspectratio]{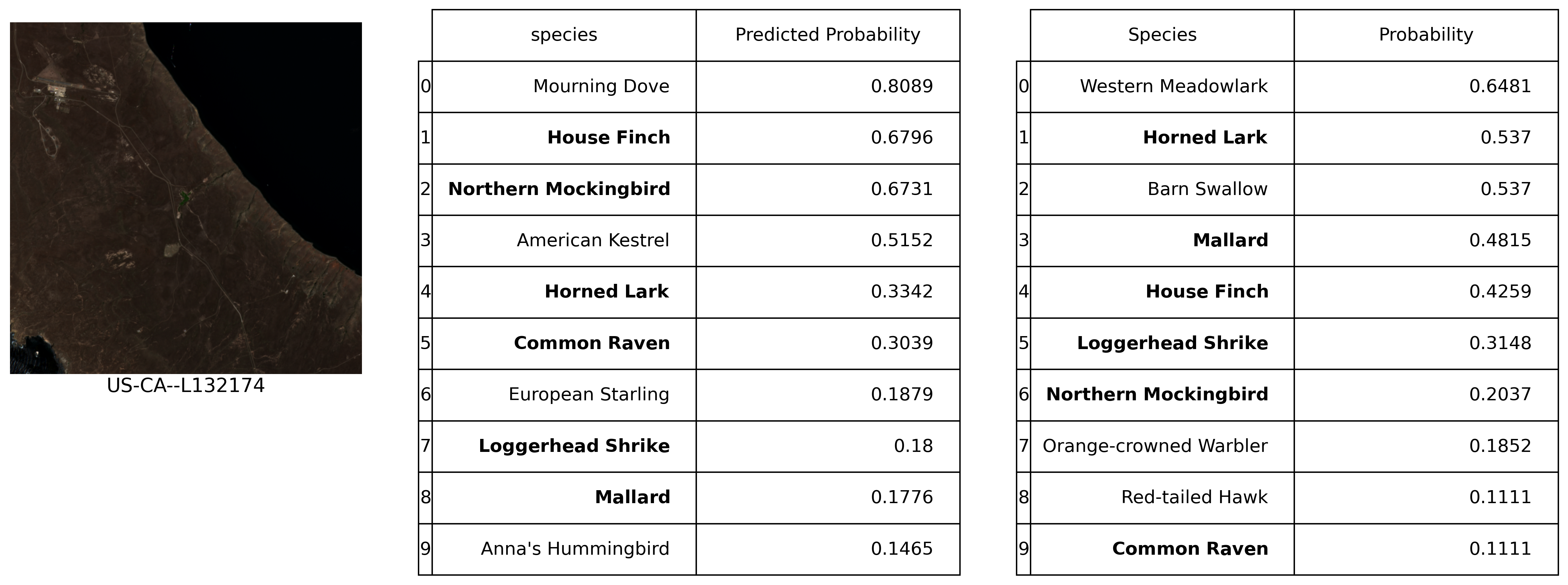}
    \caption{Hotspot and species at San Clemente Island, California: Our model predicts the presence of various species in this region, including the house finch and loggerhead shrike, which are consistent with the ground truth and highly reported on eBird.
 }
    \label{fig:hs1}
\end{figure}

\begin{figure}[h]
    \centering
    \includegraphics[width=\textwidth,height=0.9\textheight,keepaspectratio]{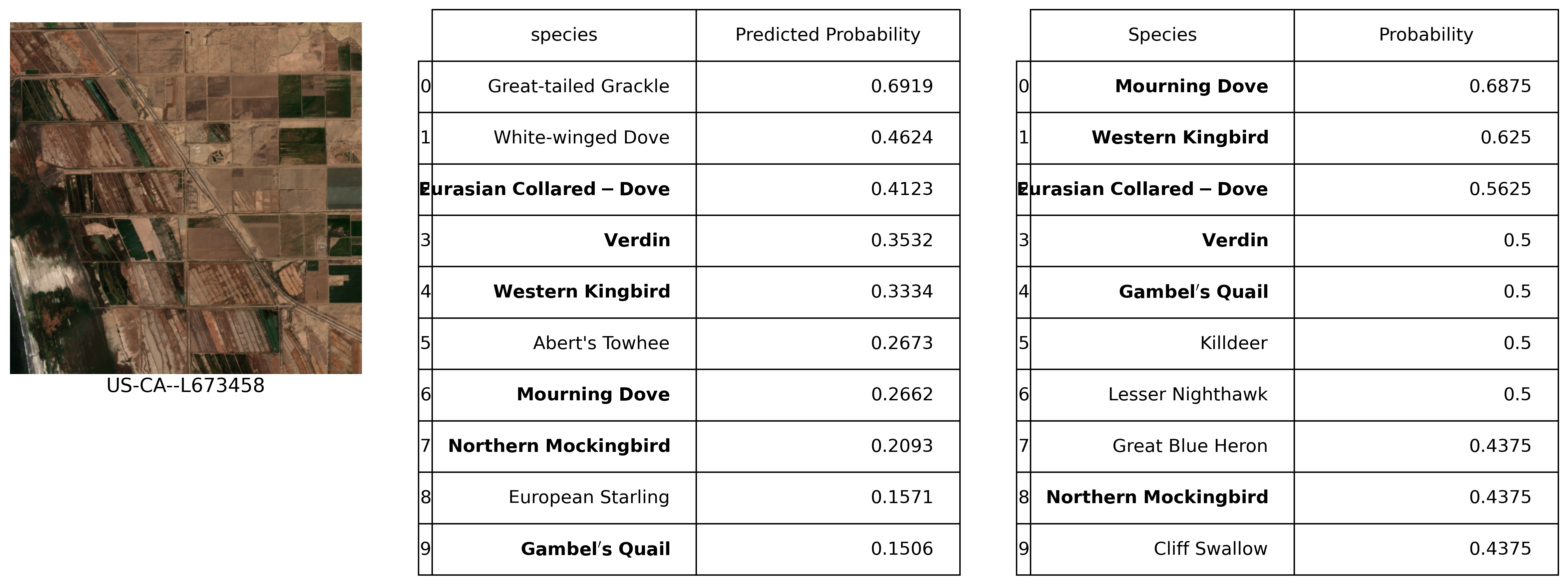}
    \caption{Hotspot and species at Salton Sea, California: Our models predict the presence of various species in this region, including the western kingbird, verdin, and Eurasian collared dove. These species are consistent with the ground truth and are highly reported on eBird for this location.}
    \label{fig:hs2}
\end{figure}

%  MAINE 

\begin{figure}[h]
    \centering
    \includegraphics[width=\textwidth,height=0.9\textheight,keepaspectratio]{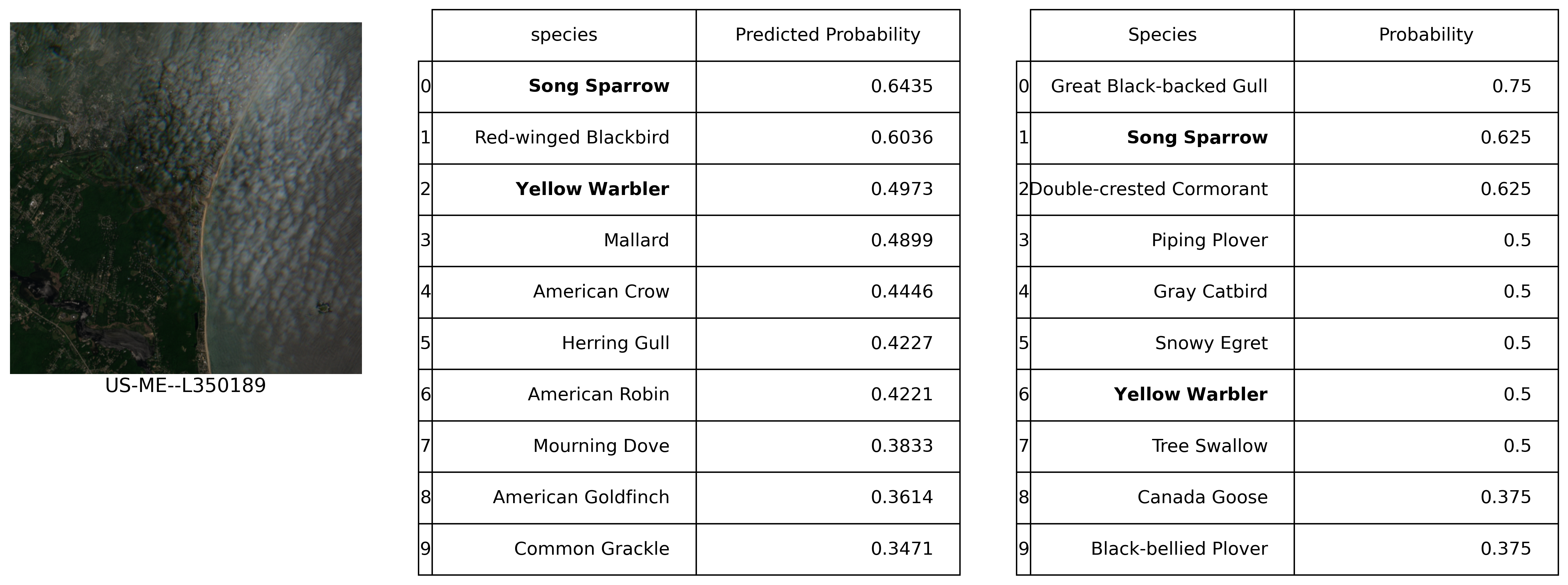}
    \caption{Hotspot and species at Rachel Carson National Wildlife Refuge, Maine: Our models correctly identified the song sparrow and the yellow warbler in this region, which exist in the top-10 species in both the ground truth and predicted list. However, it is worth noting that shoreline species are absent from the predicted top-10 list, representing a failure of our model. This may be a consequence of the partial cloud cover in the satellite image. }
    \label{fig:hs3}
\end{figure}

\begin{figure}[h]
    \centering
    \includegraphics[width=\textwidth,height=0.9\textheight,keepaspectratio]{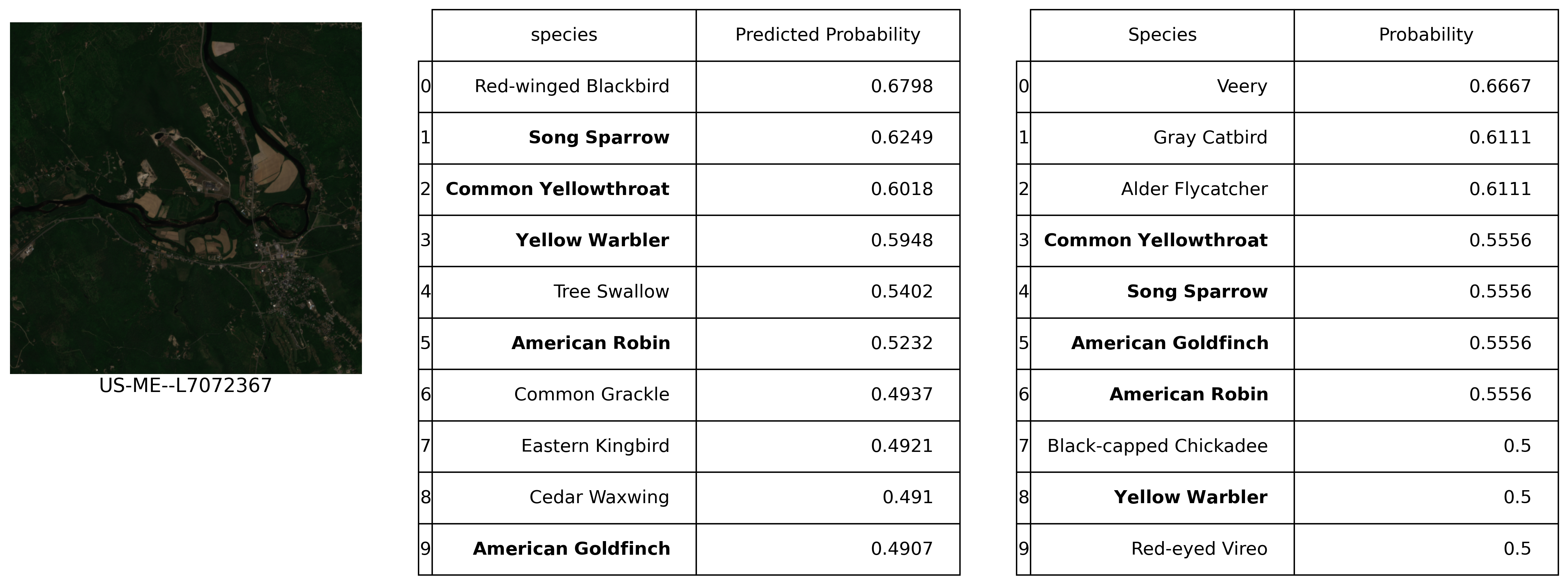}
    \caption{At Valentine Farm Conservation Center in Maine, our models correctly identified the song sparrow, common yellowthroat, yellow warbler, and American goldfinch. These species exist in both the ground truth and are consistent with Maine's ecology. 
}
    \label{fig:hs4}
\end{figure}

% OHIO ! 
\begin{figure}[!h]
    \centering
    \includegraphics[width=\textwidth,height=0.9\textheight,keepaspectratio]{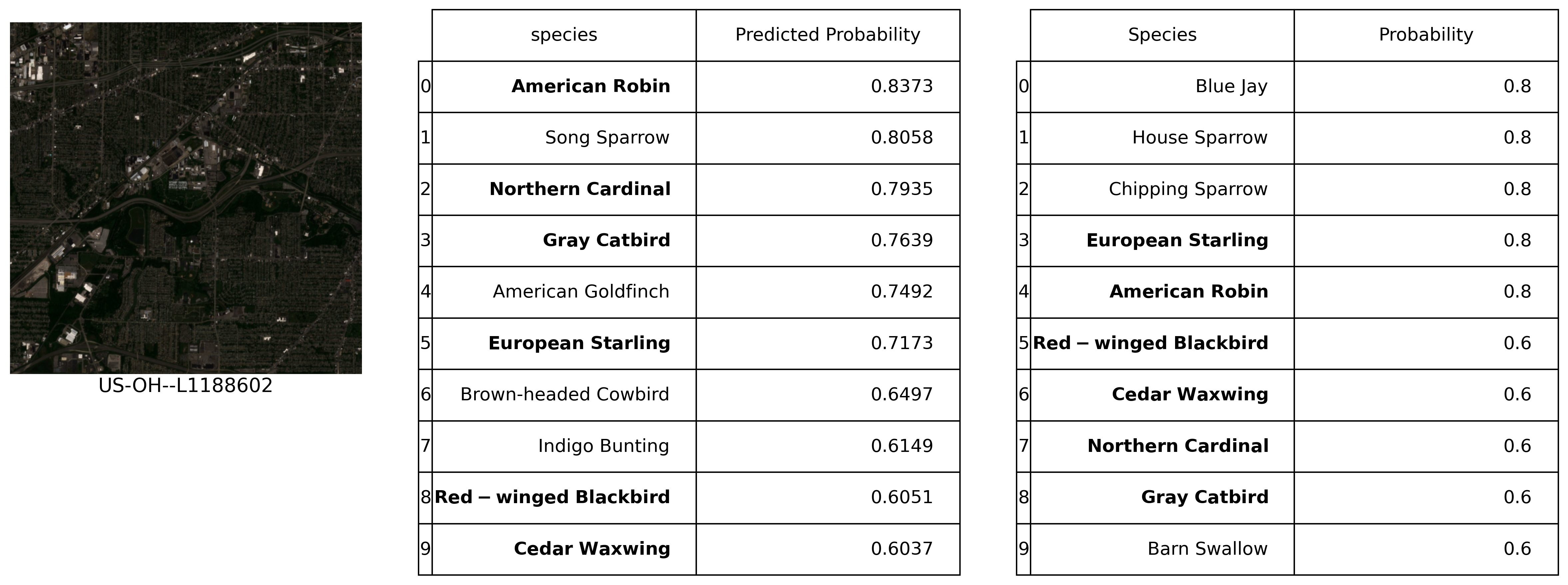}
    \caption{At West Park Cemetery in Ohio, our models accurately identified the American robin, northern cardinal, gray catbird, and six of the top ten species in this hotspot. This is consistent with the species reported on eBird.}
    \label{fig:hs5}
\end{figure}

\subsection{Analyzing predictions}\label{sup:precrec}
In addition to adaptive Top-$k$, Top-10 and Top-30 accuracy, which evaluate the model globally across species, we consider precision and recall per species to identify which are the species for which our models perform the best or worse. Our targets are continuous but to compute the precision and recall, we treat them in the same way as for as our accuracy metrics. 
In Table \ref{tab:top50}, we report the top 50 species according to recall metric for the  RGBNIR + Env + RM model. These are the species whose presence the model does not fail to predict.  
We also list some species which have 0 recall in Table \ref{tab:bot} but note that all species which have low recall are seen in very few hotspots.
\begin{table}[!h]
\centering
\small
\begin{tabular}{|l|l|l|}
\hline
\textbf{Species}      & \textbf{Recall} & \textbf{ Occurrences}  \\\hline
Long-eared Owl         & 0.0    & 4.0         \\
Pacific Golden-Plover  & 0.0    & 1.0         \\
American Golden-Plover & 0.0    & 2.0         \\
Snow Bunting           & 0.0    & 1.0         \\
Tufted Puffin          & 0.0    & 1.0         \\
Greater Scaup          & 0.0    & 9.0         \\
Black-footed Albatross & 0.0    & 1.0         \\
Red Phalarope          & 0.0    & 1.0         \\
Northern Fulmar        & 0.0    & 2.0         \\
Harris's Sparrow       & 0.0    & 1.0    \\
\hline
\end{tabular}
\caption{Example species with 0 recall on the test set,which means the model misses them but we also note that they are encountered in very few hotspots.}
\label{tab:bot}
\end{table}
\begin{table}[!h]
\centering
\small
\begin{tabular}{|l|c|c|c|}

\hline
\textbf{Species}           & \textbf{Recall} &\textbf{ Precision} &\textbf{ Occurrences} \\ \hline
Varied Bunting             & 1.000 & 1.000 & 4   \\
Botteri's Sparrow          & 1.000 & 1.000 & 5   \\
Olive Sparrow              & 1.000 & 1.000 & 4   \\
Pink-footed Shearwater     & 1.000 & 1.000 & 1   \\
Sprague's Pipit            & 1.000 & 1.000 & 1   \\
Cassia Crossbill           & 1.000 & 1.000 & 1   \\
Red-faced Warbler          & 1.000 & 0.875 & 7   \\
Black-crested Titmouse     & 1.000 & 0.857 & 6   \\
Yellow-eyed Junco          & 1.000 & 0.833 & 5   \\
Abert's Towhee             & 1.000 & 0.813 & 13  \\
Bicknell's Thrush          & 1.000 & 0.800 & 4   \\
Rufous-winged Sparrow      & 1.000 & 0.800 & 4   \\
Great Kiskadee             & 1.000 & 0.800 & 4   \\
Blue-throated Mountain-gem & 1.000 & 0.800 & 4   \\
Gambel's Quail             & 1.000 & 0.741 & 20  \\
Verdin                     & 1.000 & 0.714 & 25  \\
Reddish Egret              & 1.000 & 0.700 & 7   \\
Arizona Woodpecker         & 1.000 & 0.667 & 4   \\
Mexican Chickadee          & 1.000 & 0.667 & 2   \\
Bridled Titmouse           & 1.000 & 0.636 & 7   \\
Gray Hawk                  & 1.000 & 0.571 & 4   \\
Olive Warbler              & 1.000 & 0.500 & 3   \\
Baird's Sparrow            & 1.000 & 0.500 & 1   \\
Mexican Duck               & 1.000 & 0.500 & 2   \\
Mountain Plover            & 1.000 & 0.500 & 1   \\
Buff-breasted Flycatcher   & 1.000 & 0.500 & 2   \\
Snail Kite                 & 1.000 & 0.500 & 1   \\
Scaled Quail               & 1.000 & 0.500 & 5   \\
Wilson's Plover            & 1.000 & 0.444 & 4   \\
Nanday Parakeet            & 1.000 & 0.429 & 3   \\
California Gnatcatcher     & 1.000 & 0.400 & 2   \\
Whiskered Screech-Owl      & 1.000 & 0.400 & 2   \\
Green Kingfisher           & 1.000 & 0.333 & 1   \\
Golden-cheeked Warbler     & 1.000 & 0.333 & 1   \\
Gray-crowned Rosy-Finch    & 1.000 & 0.333 & 1   \\
Thick-billed Longspur      & 1.000 & 0.250 & 1   \\
Kirtland's Warbler         & 1.000 & 0.250 & 1   \\
Brown-capped Rosy-Finch    & 1.000 & 0.250 & 1   \\
Red-cockaded Woodpecker    & 1.000 & 0.200 & 1   \\
Northern Cardinal          & 0.985 & 0.889 & 480 \\
Mourning Dove              & 0.982 & 0.897 & 720 \\
Common Grackle             & 0.978 & 0.882 & 504 \\
Western Gull               & 0.978 & 0.786 & 45  \\
Black-billed Magpie        & 0.972 & 0.812 & 71  \\
American Robin             & 0.969 & 0.914 & 678 \\
Ladder-backed Woodpecker   & 0.968 & 0.682 & 31  \\
American Crow              & 0.966 & 0.871 & 651 \\
Carolina Chickadee         & 0.966 & 0.778 & 174 \\
Blue Jay                   & 0.961 & 0.897 & 490 \\
Carolina Wren              & 0.956 & 0.801 & 294\\
\hline
\end{tabular}
\caption{Top 50 species with highest recall on the test set, recall and precision per species scores and number of test set hotspots in which species have non-zero encounter rates, for predictions of the RGBNIR + Env + RM model. Note that only species which were encountered in the test set are considered (566/684). Rows are sorted by recall performance. }
\label{tab:top50}

\end{table}

\section{Acknowledgements}
We are grateful to Srishti Yadav for initial dataset exploration, Sal Elkafrawy for investigating the  initialization of our models, and Dan Morris and Nebojsa Jojic for insights in remote sensing. This project was supported by a Microsoft-Mila collaboration grant and a Microsoft AI for Earth cloud compute grant. Mélisande Teng received the support of the NSERC-CREATE LEADS program and Benjamin Akera received support from IVADO. David Rolnick was supported in part by the Canada CIFAR AI Chairs Program. The authors also acknowledge material support from NVIDIA in the form of computational resources, and are grateful for technical support from the Mila IDT team in maintaining the Mila Compute Cluster.

\section{Contributions}
We are happy to say that this work was a truly collaborative effort, and that this work would not have been possible without any one of us. We will just highlight a few of the many contributions each one of us has made to this project.  \\
In particular, \textbf{Mélisande Teng} played a major role in the problem and task formulation through multiple iterations since the beginning of the project and led the dataset building efforts. \textbf{Amna Elmustafa} played an instrumental role in incorporating geographical information in our models. \textbf{Benjamin Akera} contributed to the analysis of the predictions of our models with thorough inspections of the hotspots. 
 \textbf{Hugo Larochelle} provided valuable guidance, not only with insightful discussions about methodology but also with much appreciated moral support. 
 \textbf{David Rolnick} was the ideal supervisor for this project and originally came up with the idea of leveraging eBird and remote sensing data for understanding ecosystems better. 

\end{document}